\pdfoutput=1

\documentclass[11pt]{article}

\usepackage[]{acl}
\usepackage{times}
\usepackage{latexsym}
\usepackage{graphicx}
\usepackage{arydshln}
\usepackage{subfigure}
\usepackage{makecell}
\usepackage{colortbl}
\usepackage{pifont}
\usepackage{amssymb}

\usepackage[T1]{fontenc}
\usepackage{xcolor}
\usepackage[utf8]{inputenc}

\usepackage{microtype}

\usepackage{inconsolata}

\usepackage{array}
\usepackage{pifont}
\usepackage{tabularx}
\usepackage{adjustbox}
\usepackage{algorithm}
\usepackage{algpseudocode}
\usepackage{algorithmicx}
\usepackage{multirow}
\usepackage{enumitem}
\usepackage{xspace}
\usepackage{tcolorbox}
\usepackage{xparse}
\usepackage{soul}
\usepackage{booktabs,amsfonts,dcolumn}
\usepackage{hyperref}
\usepackage{url}
\usepackage{amsmath,amsthm,amsfonts,amssymb,bm,stmaryrd}
\usepackage{cleveref}

\usepackage[hang,flushmargin]{footmisc}

\newcommand{\ours}{{{REST}}}

\title{{\ours}: Retrieval-Based Speculative Decoding
}

\author{%
Zhenyu He$^{1*}$, Zexuan Zhong$^{2*}$,Tianle Cai$^{2*}$, Jason D. Lee$^{2}$, Di He$^{1\dagger}$\\ \\
$^1$National Key Lab of General AI, School of Artificial Intelligence, Peking University\\ $^2$Princeton University \\
\texttt{hezhenyu@stu.pku.edu.cn}, \quad
\texttt{zzhong@cs.princeton.edu}, \\ \texttt{\{tianle.cai, jasonlee\}@princeton.edu},\quad \texttt{dihe@pku.edu.cn}
}

\begin{document}
\maketitle
\begin{abstract}
We introduce \underline{Re}trieval-Based \underline{S}pecula\underline{t}ive Decoding ({\ours}), a novel algorithm designed to speed up language model generation. The key insight driving the development of REST is the observation that the process of text generation often includes certain common phases and patterns. Unlike previous methods that rely on a draft language model for speculative decoding, {\ours} harnesses the power of retrieval to generate draft tokens. This method draws from the reservoir of existing knowledge, retrieving and employing relevant tokens based on the current context. Its plug-and-play nature allows for seamless integration and acceleration of any language model, all without necessitating additional training. When benchmarked on 7B and 13B language models in a single-batch setting, {\ours} achieves a significant speedup of $1.62\times$ to $2.36\times$ on code or text generation. The source code of REST is available at \url{https://github.com/FasterDecoding/REST}.

\end{abstract}

\renewcommand{\thefootnote}{\fnsymbol{footnote}}
\footnotetext[1]{These three authors contributed equally to this project.}
\footnotetext[2]{Correspondence to: Di He <\texttt{dihe@pku.edu.cn}>.}
\renewcommand{\thefootnote}

\section{Introduction}


Transformer-based Large Language Models (LLMs) have emerged as a foundation model in natural language processing~\citep{vaswani2017attention, devlin2019bert, brown2020language, zhang2022opt, scao2022bloom, chowdhery2022palm, zeng2022glm, touvron2023llama}. While they achieve impressive performance across various tasks, the inference cost is huge in practical scenarios. During inference, the model autoregressively uses the preceding context to generate the next token. Each iteration requires reloading the billion-parameter LLM from the High-Bandwidth Memory (HBM) to the on-chip cache of modern accelerators like GPUs, making the whole generation inefficient and time-consuming.

A recent direction in accelerating the LLM generation is to reduce the number of forward processes with LLMs while guaranteeing the quality of the output sequence simultaneously. Speculative decoding~\citep{Leviathan2023speculative, chen2023accelerating, miao2023specinfer, spector2023accelerating} is one of the typical approaches in this direction. Intuitively, speculative decoding methods leverage a small LM to generate tokens with less computational cost. During inference, the method first uses the small LM to create a \emph{draft token sequence} and then uses the LLM for verification. If the predictions from both models are consistent, we can accept the draft and return it to the user. Here, the actual token generation is carried out using the small LM, and the large LM is only used to validate the draft, which can be performed \emph{in parallel} and requires reloading the memory only once. Consequently, the entire framework of speculative decoding reduces the overall inference cost.

\begin{figure*}[t]
\centering
\includegraphics[width=1.0\linewidth]{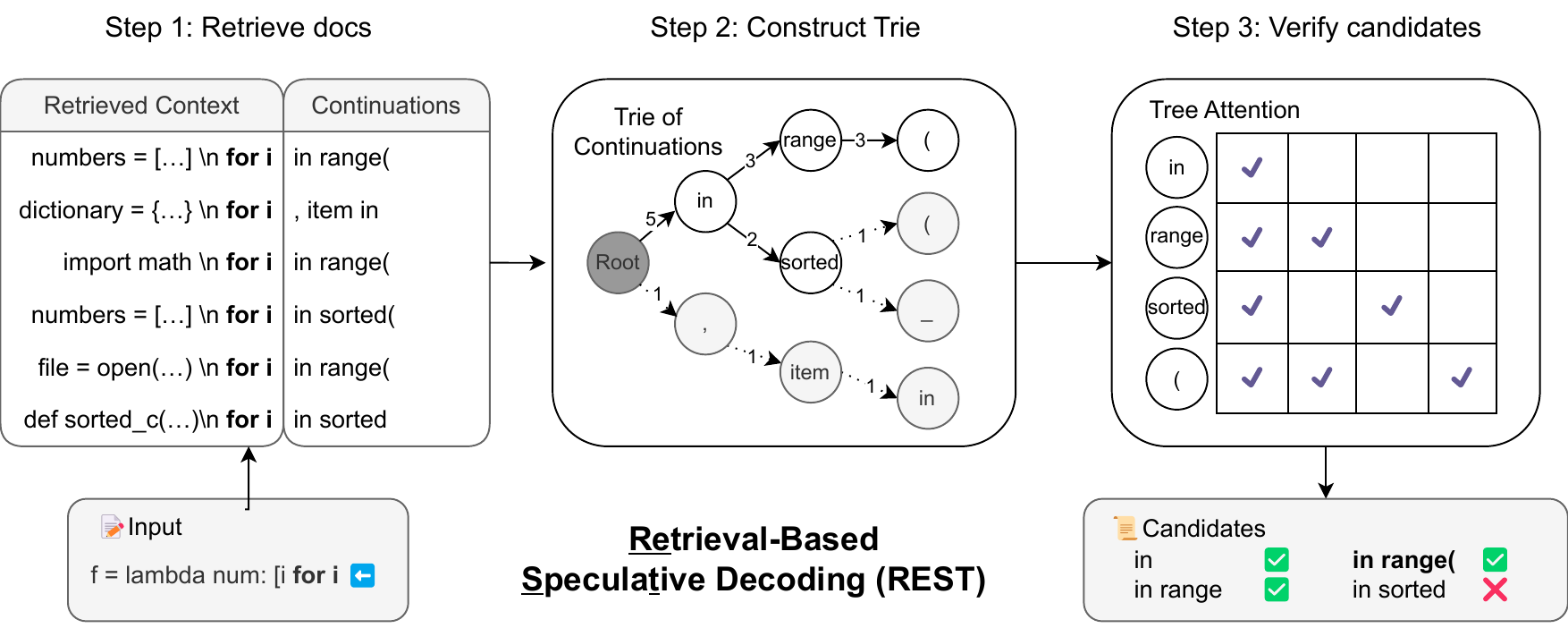}
\caption{
Overview of {\ours}. During inference, the input context is utilized as the query to retrieve docs from the datastore that match the longest suffix of the input. A Trie is constructed using the continuations from the retrieved docs. We prune the low-frequency (weight) branches and the remaining subtree is further used as candidates. Candidates will be fed into the LLM with a tree attention mask for verification. All correct tokens from the start will be accepted, and the draft tokens after the first mistake will be rejected.}
\label{fig:method}
\end{figure*}

However, obtaining a high-quality draft model remains challenging: It must balance small size and strong predictive power while matching the vocabulary of the base model; also, it should integrate well into a distributed system for serving. Therefore, people often need to train a draft model specifically for their model and use cases~\citep{chen2023accelerating,miao2023specinfer,medusa}. In this study, rather than relying on an additional small LM, we investigate using a data corpus directly to construct the draft token sequence in speculative decoding. We develop a retrieve-based approach, called \underline{Re}trieval-Based \underline{S}pecula\underline{t}ive Decoding ({\ours}) (Figure~\ref{fig:method}). Compared to previous approaches, our retrieval-based system replaces the parametric draft model with a non-parametric retrieval datastore, which can easily port to any LLM and accelerate its inference.

To use REST, the first step is constructing the datastore. In this paper, we leverage either the pretraining data corpus or the instruction-tuning data corpus to build our datastore, which serves as the source for the draft token sequence. During each inference step, we first use previous tokens (pre-generated tokens or prompt tokens) as queries to identify exact matches in the datastore. The subsequent tokens from these exact matches are considered generation candidates. A Trie is constructed using these candidates. The nodes with the highest frequencies are selected as the draft tokens. This sequence then undergoes verification by the LLM through a single forward pass, aided by a meticulously designed attention mask known as tree attention~\cite{medusa, miao2023specinfer, spector2023accelerating}. As many subsequences during generation likely appear in the datastore, REST can frequently generate multiple correct tokens per step.

We conduct extensive experiments to test the efficiency and effectiveness of REST in different scenarios. For the code domain, we use a portion of Python pretraining code (2.7M samples) from The Stack~\cite{Kocetkov2022TheStack} as the datastore and accelerate CodeLlama~\cite{rozière2023codellama} 7B and 13B respectively. The results show on HumanEval~\cite{chen2021evaluating} REST achieves $2.12\times$ to $2.36\times$ speedup. For the general domain, we construct a datastore using UltraChat~\cite{ultrachat}, containing around 774K conversations. The results show on MT-Bench~\cite{zheng2023judging} REST accelerates 7B and 13B Vicuna~\cite{chiang2023vicuna} by $1.62\times$ to $1.77\times$ respectively.

\section{Related Work}

Improving the efficiency of LLM inference has been an emergent research direction in recent years. Broadly, previous attempts can be divided into two categories: lossless acceleration and lossy acceleration.
Lossy acceleration approaches aim to learn efficient models that can execute faster and act similarly to a target LLM. These methods include pruning~\citep{wang2021spatten, hubara2021accelerated, ma2023llmpruner, frantar2023sparsegpt}, quantization~\citep{yao2022zeroquant, park2022nuqmm, dettmers2022gpt3, frantar2022optq, xiao2023smoothquant, liu2023llmqat} and knowledge distillation~\citep{sanh2019distilbert}. Lossless acceleration strategies focus on directly accelerating the target LLM from different perspectives, such as memory and IO optimization~\citep{dao2022flashattention, dao2023flashattention, kwon2023efficient, sheng2023flexgen}, and ways to reduce the function calls of LLM during decoding, e.g., speculative decoding~\citep{stern2018blockwise, Leviathan2023speculative, chen2023accelerating, miao2023specinfer, spector2023accelerating, medusa}.
This work falls within the second branch. Speculative decoding~\citep{Leviathan2023speculative, chen2023accelerating, miao2023specinfer, spector2023accelerating} leverages a smaller model to generate a draft and use LLM to verify the draft tokens with a single forward pass. In this framework, blockwise parallel decoding~\citep{stern2018blockwise} and Medusa~\citep{medusa} train multiple heads based on the LLM for draft token generation.

Our method diverges from these approaches by retrieving draft tokens from a datastore, presenting a novel avenue for efficiency improvement in large language model generation. While there is a similar study, LLMA~\citep{yang2023inference},  that employs retrieval to accelerate generation, our work distinguishes itself in two primary ways: (1) The LLMA approach is tailored towards scenarios where referred contexts (as in Retrieval-Augmented Generation and Cache-Assisted Generation) are provided during generation, and it only retrieves from these referred contexts. In contrast, our method retrieves draft tokens from a comprehensive datastore, thereby not being confined to a small context. (2) In the LLMA framework, the retrieved instance is typically limited to one or a handful. Our method, however, is designed to handle a much larger number of retrieved instances. This difference in approach allows us to leverage a wider information base during the generation process.

\section{Retrieval-Based Speculative Decoding}
In this section, we first provide notations and a background overview of speculative decoding and then introduce our proposed {\ours} framework.

\subsection{Background: Speculative Decoding}

We use $x\in \mathcal{V}$ to denote a token where $\mathcal{V}$ is the vocabulary. At each time step $t$, given the preceding context $s=(x_1, ...,x_{t-1}, x_t)$, the autoregressive decoding method generates the token at position $t+1$ according to:
\begin{align*}
x_{t+1} \sim p(x|x_1, \dots, x_t; \theta_{large}),
\end{align*}
where $p(\cdot)$ is the conditional probability distribution calculated by the LLM with parameter $\theta_{large}$. In this process, a forward run of the LLM is required at each step of generation. This is significantly time-consuming due to the memory bandwidth and cannot fully exploit the computational power of modern GPU hardware~\citep{shazeer2019fast}. 

Speculative decoding aims to reduce the computational cost during inference by reducing the count of executions with $\theta_{large}$. In addition to the LLM $\theta_{large}$, speculative decoding leverages another language model of a much smaller size with parameter $\theta_{small}$. At step $t$, the method operates by iteratively executing the following steps.

\paragraph{Draft construction}
Given $s=(x_1, \dots, x_t)$, the small LM $\theta_{small}$ is used to generate the next $m$ tokens $\tilde x_{t+1}, \dots, \tilde x_{t+m}$ in an autoregressive way: 
\begin{align*}
& \tilde x_{t+i} \sim p(x|s, \tilde x_{t+1}, \dots, \tilde x_{t+i-1}; \theta_{small}), \\
& \text{where}~~ i=1, \dots, m \nonumber.
\end{align*}
Although the tokens are still generated one by one, the computational cost of this process is reduced as it uses  $\theta_{small}$ instead of $\theta_{large}$.

\paragraph{Draft verification}
After the draft tokens $\tilde x_{t+1}, \dots, \tilde x_{t+m}$ are generated, they are fed into the LLM together with the context $s$. The LLM $\theta_{large}$ then calculates the conditional probabilities with a single forward pass:
\begin{align*}
    & p(x|s; \theta_{large}),\nonumber\\
& p(x|s, \tilde x_{t+1}; \theta_{large}), \nonumber \\
& \dots \nonumber\\
& p(x|s, \tilde x_{t+1}, \dots, \tilde x_{t+m-1}; \theta_{large}).
\end{align*}
\paragraph{Draft acceptance}
Starting from the first generated tokens in the draft, the conditional probability derived from $\theta_{small}$ is compared with that of $\theta_{large}$. We can use modified rejection sampling to match the generated distribution with the LLM~\citep{Leviathan2023speculative,chen2023accelerating}. For position $t+i$, we first sample a value $r$ from a uniform distribution $U(0, 1)$. If $r <\min\left(1, \frac{p(x|s,\tilde x_{t+1}, \dots, \tilde x_{t+i-1};\theta_{large})}{p(x|s,\tilde x_{t+1}, \dots, \tilde x_{t+i-1};\theta_{small})}\right)$, we accept the draft token $\tilde x_{t+i}$ and continue to validate the next token $\tilde x_{t+i+1}$. Otherwise, we stop the acceptance process, resample $x_{t+i}$, reject all the draft tokens after $x_{t+i}$, and move to the new speculative decoding process at the position $t+i+1$. 

\subsection{Our Approach: {\ours}}
While in the classic speculative decoding, a smaller LM is used as the draft model, finding a high-quality draft model is usually challenging for several reasons: (1) For efficiency, the draft model needs to be lightweight enough to not introduce much overhead. (2) For quality, it needs to predict the LLM output accurately. (3) For system integration, it needs the same vocabulary set as the LLM, and an architecture that distributes easily with a similar configuration to the LLM~\cite{chen2023accelerating}. These challenges require carefully selecting or even training custom draft models for each new LLM.

In this paper, we solve the challenges differently. We develop a training-free approach to speculative decoding that can easily integrate with any new model to accelerate inference. Instead of relying on a parametric draft model, our method \underline{Re}trieval-Based \underline{S}pecula\underline{t}ive Decoding ({\ours}) proposes using retrieval for draft construction. An overview of {\ours} is shown in Figure~\ref{fig:method}. In the following, we first describe constructing a datastore and operations on it, then demonstrate using it for draft construction and verification. Together, {\ours} provides an efficient, high-quality, and easy-to-integrate solution for accelerating the inference of LLMs.



\paragraph{Datastore construction}
{\ours} operates based on a pre-built datastore $D = \{(c_i, t_i)\}$, where $c_i$ represents a context and $t_i$ represents the corresponding continuation of the context $c_i$.
Given a text/code corpus, we construct the datastore $D$ using the prefix context and the corresponding continuation at each position.

\paragraph{Retrieving from the datastore}
At inference, given a context $s=(x_1, \dots, x_t)$, our objective is to construct the draft tokens which are likely the continuations of $s$.
Different from vanilla speculative decoding that uses a small LM to construct the draft, we leverage the built datastore $D$ and directly retrieve draft tokens from the datastore.
We first use the context $s$ to retrieve context-continuation pairs from the datastore $D$ and construct a set of continuation candidates $S$:
\begin{align*}
S=\{t_i \mid (c_i, t_i) \in \texttt{Retrieve}(D,s)\},
\end{align*}
where $\texttt{Retrieve}(D,s)$ implements a retrieval process in the datastore $D$ that returns a set of context-continuation pairs $\{(c_i, t_i)\}$ by using $s$ as the query.
It is straightforward to use recent dense retrieval models~\cite{khandelwal2019generalization, karpukhin2020dense} to find contexts $c_i$ that are similar to $s$.
However, using dense retrievers adds additional overhead during inference. 
We instead use a fast exact-match method to retrieve continuation candidates.

Our retrieval process is shown in Algorithm~\ref{alg:retrieval}.
We aim to find contexts in $D$ that match the longest suffix of $s$.
We employ a greedy strategy and start from a pre-defined match length upper limit $n_{max}$\footnote{We set $n_{max}$ as 16 in our experiments, as only few cases lead to a maximum match with more than $16$ tokens.}.
For each suffix length $n$, we obtain the context $s$'s suffix with $n$ tokens $q$ (line 5), and obtain all the contexts $c_i$ that match $q$ as a suffix (line 6).
If at least one context in $D$ matches the current $q$ (i.e., $S \neq \emptyset$), we return the corresponding context-continuation pairs as the retrieval result; otherwise we decrease the matching length $n$ by one and try to match a shorter suffix (line 7).
We use a suffix array~\cite{manber1993suffix} to implement efficient exact match in datastore $D$ for a given $q$. 
The retrieval process leads to negligible overhead ($<6\%$) in our experiments (see details in Section~\ref{sec:analysis}).

\begin{algorithm}[th]
\caption{Exact-match based retrieval algorithm \texttt{Retrieve}$(D,s)$. We return context-continuation pairs in $D$ that match the longest suffix of $s$.}
\label{alg:retrieval}
\begin{algorithmic}[1]
\State \textbf{Input:} Context $s$, datastore $D$, maximum suffix length $n_{max}$
\State Initialize $n \gets n_{max}$
\State Initialize $S \gets \emptyset$ 
\While{$S = \emptyset$}
    \State $q \gets \texttt{suffix}(s, n)$
    \State $S \gets \{(c_i, t_i) \mid q = \texttt{suffix}(c_i, n)\}\subseteq D$
    \State $n \gets n - 1$
\EndWhile
    \State \textbf{return} $S$
\end{algorithmic}
\end{algorithm}

\paragraph{Draft construction from retrieved results}
The retrieved result $S$ includes possible continuations of the context $s$. 
For each $t_i \in S$, any prefix of $t_i$ can serve as draft tokens of $s$ in the speculative decoding and be further verified by the LLM.
Note that the retrieved set of continuation candidates $S$ can be large.
It is not feasible to use all candidates as draft tokens and feed them into the LLM for verification.
Here we present how we select high-quality draft tokens from the retrieved set $S$.
A naive strategy is to sample a subset of sequences in $S$ as the draft tokens. However, this is suboptimal as the sampling contains randomness when $S$ is large. 

We select draft tokens from the retrieved result $S$ using a Trie. In the Trie, the unique path from a node to the root node corresponds to a prefix of $t_i \in S$. For each node, we assign a weight reflecting the number (frequency) of the corresponding prefix that appears in the retrieved candidates. As shown in Algorithm~\ref{alg:trie}, we first construct a Trie using all sequences in $S$, and the node weight is updated when a candidate $t_i$ is inserted into the Trie (lines 2-7). The Trie data structure allows us to prioritize tokens using the weights and select high-frequency prefixes (lines 8-15). In the practical implementation, we choose a subtree that contains the top $c$ nodes with the highest weights, which equals to selecting the top $c$ high-frequency prefixes as the draft sequences.

\begin{algorithm}
\caption{Draft sequences selection using Trie.}\label{alg:trie}
\begin{algorithmic}[1]
\State \textbf{Input:}Continuation Candidates $S$, hyperparameter $c$
\State Initialize Trie $T$
\For{each $t_i \in S$}
    \For {each $prefix$ of $t_i$}
        \State Insert $prefix$ into $T$ and update node weights
    \EndFor
\EndFor
\State Initialize empty priority queue $Q$ (Max Heap based on node weights)
\For{each $node$ in $T$}
\State Add $(node.prefix, node.weight)$ to $Q$
\EndFor
\While{$Q.size > c$}
    \State Pop the $prefix$ with the smallest weight from $Q$
\EndWhile
\State \textbf{return} $Q$
\end{algorithmic}
\end{algorithm}


\paragraph{Draft verification of {\ours}}
In {\ours}, multiple draft sequences may be retrieved from the datastore. While one might initially approach the drafts independently and feed them into the LLM as distinct sequences in a batch, practical observations reveal that many drafts share common prefixes. This leads to redundant computation of Transformer layers on these shared prefixes across different sequences, resulting in a waste of computational power. To optimize the efficiency, we construct a pseudo sequence from the subtree using breadth-first search. By definition, it can be immediately obtained that each draft constitutes a sub-sequence of this pseudo sequence, and any shared prefix appears only once. To correctly execute LLM on this pseudo sequence, we implement a carefully designed attention mask in each attention layer, ensuring that the computation of each token precisely reflects its dependencies in the original draft sequence. This attention strategy is also known as tree attention~\citep{medusa, miao2023specinfer, spector2023accelerating}.

\paragraph{Draft acceptance of {\ours}}
We adopt a more straightforward acceptance strategy compared to the original speculative decoding. By feeding the drafts into LLM, we obtain the conditional distribution at each position given by $\theta_{large}$, where we sample new tokens. We then assess whether sampled new tokens coincide with the draft tokens at each position. All correct draft tokens from the start will be accepted, and the draft tokens after the first mistake will be rejected. In this way, the sequences produced using {\ours} are identical to those generated by standard autoregressive generation.


\paragraph{Comparison with existing approaches}
Although {\ours} follows a schema similar to that of speculative decoding, it offers significant advantages over existing approaches.
Current speculative decoding methods rely on a high-quality small model to generate draft tokens~\cite{Leviathan2023speculative, chen2023accelerating}. 
Such methods must strike a balance between a small size and strong predictive power, while also matching the vocabulary of the base model. 
Moreover, they require additional GPU memory and introduce complexity during inference.
In contrast, {\ours} directly retrieves draft tokens from a datastore, which can be easily integrated with language models of any size, vocabulary, or architecture.
Different from \citet{stern2018blockwise} and \citet{medusa} which train specialized modules to create a draft model, {\ours} eliminates the need for any additional training steps and can serve as a plug-and-play solution of efficient decoding across different models.
Furthermore, the effectiveness of {\ours} is affected by the quality of retrieval results. This opens up the opportunities to further enhance {\ours} by using a better/larger datastore or an advanced retrieval model. We also note that in addition to using {\ours} directly, it is possible to combine {\ours} with the vanilla speculative decoding. This combination can enhance the generation speed of the small LM. We leave this for future work.

\section{Experiments}

\subsection{Experimental Setup}

\begin{table*}[ht]
    \centering
    {
    \resizebox{2.05\columnwidth}{!}{
    \vspace{.5em}
    \begin{tabular}{lccccc}
    \toprule
    \textbf{Benchmark} & \textbf{Model} & \textbf{Method} & \textbf{\textit{Mean Token Time}}($\downarrow$) & \textbf{Speedup}($\uparrow$) \\
     \midrule
    \midrule
     \multirow{6}{*}{HumanEval (1 shot)} & CodeLlama 7B & Autoregressive (Greedy) & 27.89 ms/token & $1\times$\\
    & CodeLlama 7B & Speculative (Greedy) & 15.90 ms/token & $1.75\times$\\
       &CodeLlama 7B  & REST (Greedy)  &  11.82 ms/token  & $2.36 \times$\\
     & CodeLlama 13B & Autoregressive (Greedy) & 44.32 ms/token & $1\times$\\
     & CodeLlama 13B & Speculative (Greedy) & 19.39 ms/token & $2.29\times$\\
     & CodeLlama 13B & REST (Greedy) & 19.53 ms/token & $2.27 \times$\\

    \midrule
     \multirow{6}{*}{HumanEval (10 shot)} & CodeLlama 7B & Autoregressive (Nucleus) & 27.99 ms/token & $1\times$\\
    & CodeLlama 7B & Speculative (Nucleus) & 18.83 ms/token & $1.49\times$\\
       & CodeLlama 7B  & REST (Nucleus) & 13.18 ms/token & $2.12 \times$\\
     & CodeLlama 13B & Autoregressive (Nucleus) & 44.46 ms/token & $1\times$\\
     & CodeLlama 13B & Speculative (Nucleus) & 22.68 ms/token & $1.96\times$\\
     & CodeLlama 13B & REST (Nucleus) & 20.47 ms/token & $2.17 \times$\\
     \midrule
    \midrule
     \multirow{6}{*}{MT-Bench} & Vicuna 7B & Autoregressive (Greedy) & 25.48 ms/token & $1\times$\\
       & Vicuna 7B  & Speculative (Greedy) & 19.44 ms/token & $1.31 \times$\\
       & Vicuna 7B  & REST (Greedy) & 15.12 ms/token & $1.69 \times$\\
     & Vicuna 13B & Autoregressive (Greedy) & 44.30 ms/token & $1\times$\\
     & Vicuna 13B & Speculative (Greedy) & 29.80 ms/token & $1.49\times$\\
     & Vicuna 13B & REST (Greedy) & 25.08 ms/token & $1.77 \times$\\
    \midrule
     \multirow{6}{*}{MT-Bench} & Vicuna 7B & Autoregressive (Nucleus) & 25.93 ms/token & $1\times$\\
       & Vicuna 7B & Speculative(Nucleus) & 20.65 ms/token & $1.26 \times$\\
       & Vicuna 7B & REST(Nucleus) & 16.02 ms/token & $1.62 \times$\\
     & Vicuna 13B & Autoregressive (Nucleus) & 44.32 ms/token & $1\times$\\
     & Vicuna 13B & Speculative (Nucleus) & 31.78 ms/token & $1.39\times$\\
     & Vicuna 13B & REST (Nucleus) & 25.92 ms/token & $1.71 \times$\\
    \bottomrule
    \end{tabular}
    }
    }
    \caption{Speed on HumanEval and MT-Bench with standard autoregressive decoding, speculative decoding and {\ours}. The temperature is set to 0.8 and the top-$p$ to 0.95 for nucleus sampling in HumanEval. For MT-Bench, the settings are 0.7 for temperature and 0.8 for top-$p$. For speculative decoding, we conduct experiments using different numbers of draft tokens and different small LMs and record the best results (detailed results can be found in Appendix~\ref{app:speculative}). All the experiments are conducted on a single NVIDIA A6000 GPU and 96 CPU cores with a batch size of 1. } 
    \label{tab:4_main_results}
\end{table*}

\paragraph{Sampling strategies} We implement two sampling mechanisms: greedy sampling and nucleus sampling~\citep{holtzman2019curious} for the LLM. Greedy sampling selects the token with the highest probability at each step. Nucleus sampling, also known as top-$p$ sampling, generates tokens by sampling from the most probable tokens in the model's predicted distribution until their cumulative probability reaches the threshold $p$. It is worth noting that under our approach, we only accept draft tokens if they match the tokens sampled from the LLM. As a result, the sequences produced using {\ours} are identical to those generated by standard autoregressive generation.

\paragraph{Datasets and models} We conduct experiments on two datasets: HumanEval~\citep{chen2021evaluating} and MT-Bench~\citep{zheng2023judging}. HumanEval is a dataset that includes 164 human-written Python programming problems. The goal for the models is to generate code solutions using provided docstrings as prompts. On the other hand, MT-Bench contains 80 multi-turn questions designed to emulate real-world multi-turn dialogues. We compare the generation speed of standard autoregressive generation with {\ours}, focusing on both the HumanEval and MT-Bench datasets. For HumanEval, we perform 1-shot evaluation for greedy sampling and 10-shot evaluation for nucleus sampling and employ the CodeLlama~\citep{rozière2023codellama}. While for MT-Bench, we perform 1-shot evaluation for both greedy sampling and nucleus sampling and utilize Vicuna~\citep{chiang2023vicuna}.  We test both the 7B and 13B configurations of CodeLlama and Vicuna, with a maximum generation limit of 512 tokens and 1024 tokens, respectively.  All experiments are conducted on a single NVIDIA A6000 GPU and 96 CPU cores. All results are averaged across three different runs.

\paragraph{Hyperparameters} When performing exact match in the datastore, the starting context suffix length, $n_{max}$, is set to 16, and is progressively reduced by one until we find matching contexts in the datastore. The length of each retrieved continuation candidate denoted as $m$, is truncated to 10. Empirical results from Medusa~\citep{medusa} suggest 64 draft tokens to be an optimal computation configuration. Hence, we limit the maximum number of selected draft tokens in the constructed Trie to 64, designated as $c$.

\paragraph{Metrics} The first metric we use is \textit{Mean Token Time}, which is the average generation time of one token for the LLM. Another metric, \textit{Mean Generated Length}, is calculated as the ratio of the length of the generated tokens to the number of forward steps taken by the original LLM. Formally, if $L$ denotes the length of the generated tokens and $F$ represents the number of forward steps, the \textit{Mean Generated Length}, $M$, is given by:
\begin{equation*}
M = \frac{L}{F}.
\end{equation*}
Note that the \textit{Mean Generated Length} ($M$) acts as the upper limit of the speedup that {\ours} can achieve, ignoring the overhead for retrieving and constructing draft tokens.

\paragraph{Datastores} For CodeLlama, we construct a datastore using a portion of the Python pretraining code from The Stack~\citep{Kocetkov2022TheStack}. This dataset comprises approximately 2.7M Python code samples and results in a datastore with a size of 27GB. On the other hand, for Vicuna, we construct a datastore using data derived from UltraChat~\cite{ultrachat}. This dataset consists of around 774K conversations from ChatGPT\footnote{https://chat.openai.com/}, yielding a datastore with a size of 12GB.

\paragraph{Baseline} We implement speculative decoding~\citep{Leviathan2023speculative, chen2023accelerating} as the baseline for comparison. For the small draft LMs, we test a variety of model sizes, including Llama 68M and Llama 160M trained by ~\citet{miao2023specinfer}, TinyLlama 1.1B and TinyLlama-Chat 1.1B trained by~\citet{tinyllama}. We also test different numbers of draft tokens ranging from 1 to 15 (performance degrades when larger than 15).

\subsection{Main Results} \label{sec:main_results}
Table~\ref{tab:4_main_results} compares the generation speed of {\ours} with the speed of the standard autoregressive decoding and speculative decoding.

Regarding generation speed, {\ours} demonstrates a significant speed enhancement compared to standard autoregressive decoding and speculative decoding, achieving $2.16\times$ to $2.36\times$ increase for CodeLlama in the HumanEval benchmark. The MT-Bench benchmark also reveals a speedup for Vicuna when using our method, with a factor ranging from $1.62\times$ to $1.77\times$. These empirical results lend weight to the effectiveness of our method for speeding up the generation process of LLMs. Note that the speedup of nucleus sampling is not as good as that of greedy sampling. We speculate that this drop in performance is caused by the randomness introduced by nucleus sampling. Since speculative decoding may achieve better results with a more powerful draft LM that aligns with the LLM, we do not claim that {\ours} can outperform speculative decoding under all circumstances. Yet, {\ours} undoubtedly provides a potent and straightforward approach for faster inference of LLMs.

Another intriguing observation that emerges from these results is the domain-dependent nature of the speed improvements. This characteristic has also been noted in other methods like speculative decoding~\cite{chen2023accelerating} and Medusa~\cite{medusa}. Specifically, the speedup achieved with {\ours} is significantly greater in the HumanEval benchmark than in the MT-Bench benchmark, suggesting that the effectiveness of {\ours} may vary depending on the specific domain.

Additionally, it is important to note that the average time (divided by the total number of tokens) required for retrieval (which includes the time taken to construct the Trie) is less than 1 ms. This time is very small and can, for all practical purposes, be considered negligible. This negligible retrieval time further underscores the efficiency of {\ours}.

\section{Ablation Study}
\label{sec:analysis}
\begin{table*}[t]
    \centering
    {
    \resizebox{1.9\columnwidth}{!}{
    \vspace{.5em}
    \begin{tabular}{cccccc}
    \toprule
    \textbf{Method} & \textbf{Datastore Size} & \textbf{Retrieval Time} & \textbf{\textit{M}}& \textbf{\textit{Mean Token Time}}($\downarrow$) & \textbf{Speedup}($\uparrow$) \\
     \midrule
    \midrule
     Baseline(Greedy) & - & - & 1 & 27.89 ms/token & $1\times$\\
     \midrule
       REST(Greedy) & 0.9 GB & 0.2 ms & 1.96 &  15.28 ms/token & $1.83 \times$\\
       REST(Greedy) & 4.4 GB & 0.5 ms & 2.18 & 13.98 ms/token & $1.99 \times$\\
       REST(Greedy) & 8.7 GB & 0.6 ms & 2.35 & 13.24 ms/token & $2.11 \times$\\
       REST(Greedy) & 14 GB & 0.6 ms & 2.45 & 12.99 ms/token & $2.15 \times$\\
       REST(Greedy) & 27 GB & 0.7 ms & 2.65 & 11.82 ms/token & $2.36 \times$\\
    \bottomrule
    \end{tabular}
    }
    }
    \caption{Generation speed with different datastore sizes (CodeLlama 7B with greedy sampling on HumanEval). The datastores are all constructed from the Python pretraining code from the Stack~\cite{Kocetkov2022TheStack}.} 
    \label{tab:4_abalation_datastore}
\end{table*}

\begin{figure}[t]
\centering
\includegraphics[width=1.0\linewidth]{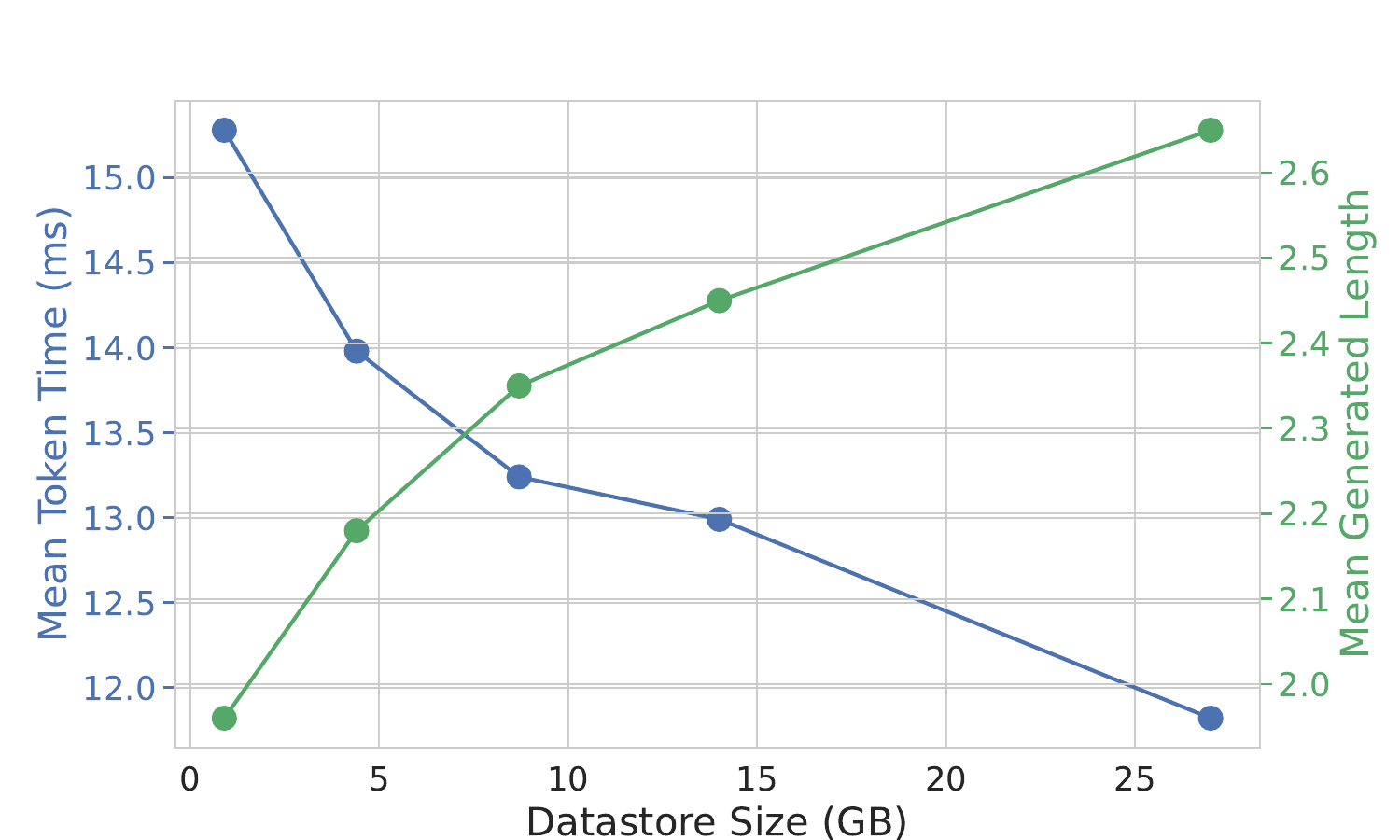}
\vspace{-0.5em}
\caption{
Generation speed of {\ours} with different sizes of the datastore (CodeLlama 7B on HumanEval). 
}
\label{fig:4_ablation_datastore}
\end{figure}

To gain a deeper understanding of our method, we conduct a series of ablation studies and analyses focused on each individual component. More ablation studies can be found in Appendix~\ref{app:ablation}. 

\paragraph{Effect of the datastore size} Increasing the size of the datastore is an effective strategy for enhancing the accuracy of retrieved draft tokens in the Trie, which in turn can significantly boost generation speed. In Table~\ref{tab:4_abalation_datastore}, we show that as the datastore size increases, both the \textit{Mean Generated Length} and \textit{Mean Token Time} correspondingly improve. However, it's important to note that the speedup growth is not as pronounced as that of the \textit{Mean Generated Length}. This discrepancy could be attributed to the overhead of getting draft tokens. We assume that in industry applications, there will be ample disk storage to build a large datastore and ample CPU cores for fast retrieval. We also visualize the trend of scaling the retrieval datastore size in Figure~\ref{fig:4_ablation_datastore}. From this, we can infer that there is still potential to achieve even faster speeds with a larger datastore.



\begin{table}[t]
    \centering
    \resizebox{1.0\columnwidth}{!}{
    \begin{tabular}{lcc}
    \toprule
    \textbf{Selecting Methods} & \textbf{\textit{M}}($\uparrow$) & \textbf{\textit{Mean Token Time}} ($\downarrow$) \\
    \midrule
    Random(Greedy) & 2.51 & 12.80 \\
    Trie(Greedy) & \textbf{2.65} & \textbf{11.82} \\
    Random(Nucleus) & 2.44 & 14.19 \\
    Trie(Nucleus) & \textbf{2.57} & \textbf{13.18} \\
    \bottomrule
    \end{tabular}
    }
    \vspace{-0.5em}
    \caption{Generation speed with different selecting methods of draft tokens (CodeLlama 7B with greedy sampling on HumanEval).}
\vspace{-0.5em}
    \label{tab:4_ablation_sample}
\end{table}

\paragraph{Effect of draft token selecting strategies} We compare selecting draft tokens in the Trie with randomly sampling retrieved continuation candidates as draft tokens. For an equitable comparison, we employ a random sampling technique to sample at most eight sequences from all the retrieved candidates. Furthermore, each sequence is truncated to a maximum length of 8. This results in a maximum number of 64 draft tokens, corresponding to the maximum number of selected draft tokens from the Trie. The data presented in Table~\ref{tab:4_ablation_sample} indicates that selecting draft tokens from the Trie, as opposed to employing a random sampling approach, enhances the performance.



\begin{figure}[t]
\centering
\includegraphics[width=1.0\linewidth]{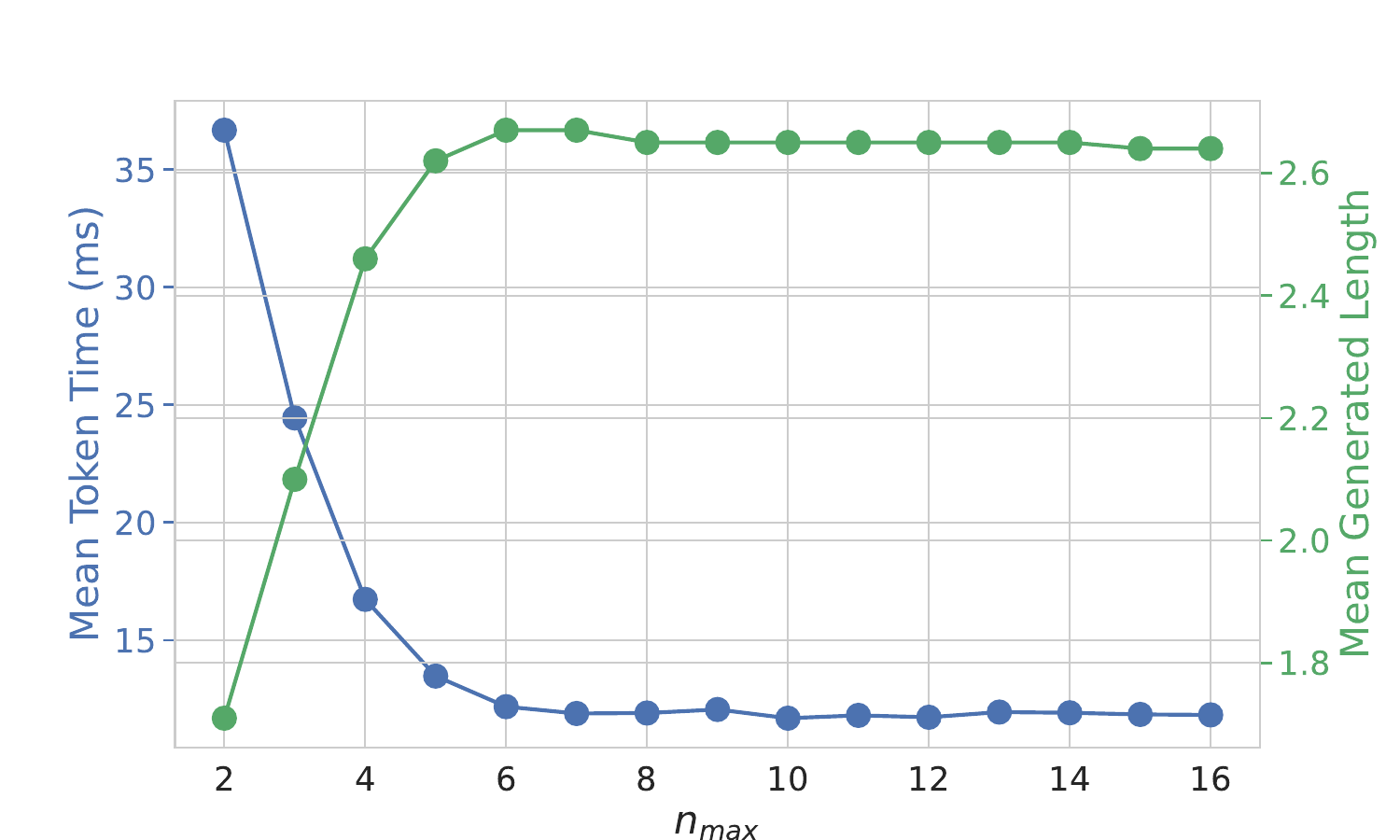}
\vspace{-0.5em}
\caption{
Generation speed of {\ours} with different maximum suffix length $n_{max}$ (CodeLlama 7B with greedy sampling on HumanEval). 
}
\label{fig:4_ablation_span}
\end{figure}

\paragraph{Effect of the choice of the maximum suffix length}
 We vary the value of $n_{max}$ to test the generation speed of {\ours}. The outcomes of this study are depicted in Figure~\ref{fig:4_ablation_span}. An interesting observation is that when the value of $n_{max}$ is set to less than 6, there is a substantial increase in the generation time. Conversely, when $n_{max}$ exceeds 6, the generation speed remains consistently high and appears to be largely unaffected by further changes to the $n_{max}$ value. Hence, in practice, there is no substantial need to expend excessive efforts in selecting the precise optimal value of $n_{max}$.

\section{Conclusion}
In this work, we propose REST: retrieval-based speculative decoding. Instead of requiring a small LM, REST employs a datastore for retrieving and employing draft tokens. We construct a Trie to select the most probable draft tokens. REST is not only straightforward to implement but also easily integrates into the generation processes of any existing language models without necessitating additional training. We would like to explore large-scale retrieval in the next step. For situations where disk storage is limited, we will also explore methods of minimizing the size of the datastore without compromising performance.

\section*{Limitations}
The limitations of our work are as follows:
\begin{itemize}
    \item Despite the plug-and-play nature of {\ours}, it is important to acknowledge that the performance of {\ours} is directly influenced by the accuracy and completeness of the datastore. For improved alignment with the LLM, it might be advantageous to consider constructing datastores from content generated by the LLM itself.
    \item Lack of in-context abilities. For instance, the challenge of retrieving personalized variable names in code generation—a task that inherently requires understanding context—raises an interesting question: How can we empower retrieval methodologies to effectively deal with such complexities?
\end{itemize}

\section*{Acknowledgement}
JDL acknowledges support of the ARO under MURI Award W911NF-11-1-0304,  the Sloan Research Fellowship, NSF CCF 2002272, NSF IIS 2107304,  NSF CIF 2212262, ONR Young Investigator Award, and NSF CAREER Award 2144994. We thank all the anonymous reviewers for the very careful and detailed reviews as well as the valuable suggestions. Their help has further enhanced our work.

\bibliography{ref}

\appendix

\newpage
\section{Detailed Results of Speculative Decoding} \label{app:speculative}
For the small draft LMs, we test a variety of model sizes, including Llama 68M and Llama 160M trained by ~\citet{miao2023specinfer}, TinyLlama 1.1B\footnote{The TinyLlama-1.1B-intermediate-step-955k-2T version.} and TinyLlama-Chat 1.1B\footnote{The TinyLlama-1.1B-Chat-V0.4 version.} trained by~\citet{tinyllama}. We also test different numbers of draft tokens from 1 to 15. Since \citet{Leviathan2023speculative} and \citet{chen2023accelerating} didn't release their code, we use an open-source reproduction code\footnote{\url{https://github.com/feifeibear/LLMSpeculativeSampling}} and additionally use \texttt{torch.compile} to accelerate the generation speed of the small draft LMs.

\subsection{Results on CodeLlama 7B}
\begin{figure*}[t]
\centering
\begin{minipage}[b]{0.49\linewidth}
  \centering
  \includegraphics[width=\linewidth]{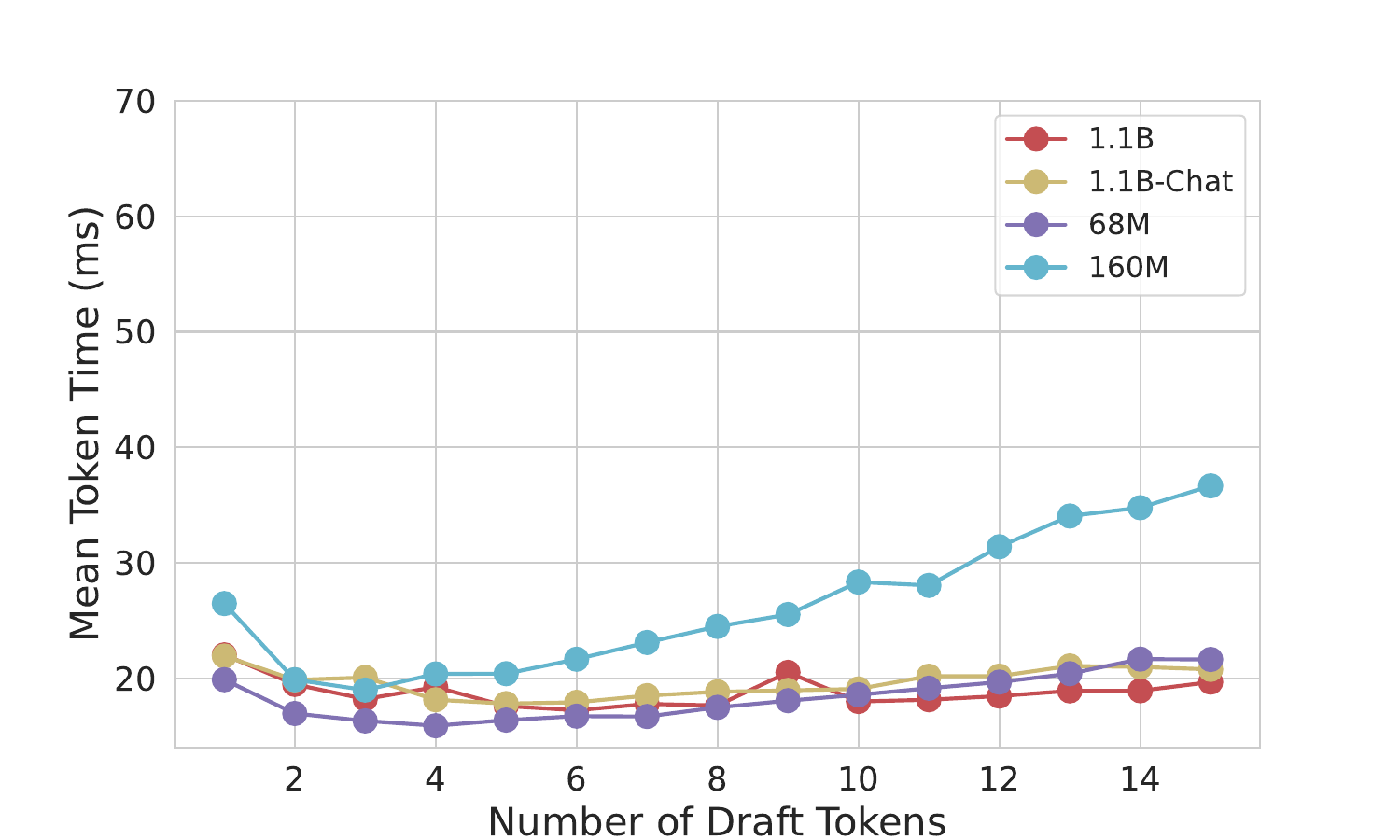}
  \caption{Generation speed of speculative decoding with different draft LMs and numbers of draft tokens (CodeLlama 7B with greedy sampling on HumanEval).}
  \label{fig:app_spec_codellama7b_greedy}
\end{minipage}
\hfill
\begin{minipage}[b]{0.49\linewidth}
  \centering
  \includegraphics[width=\linewidth]{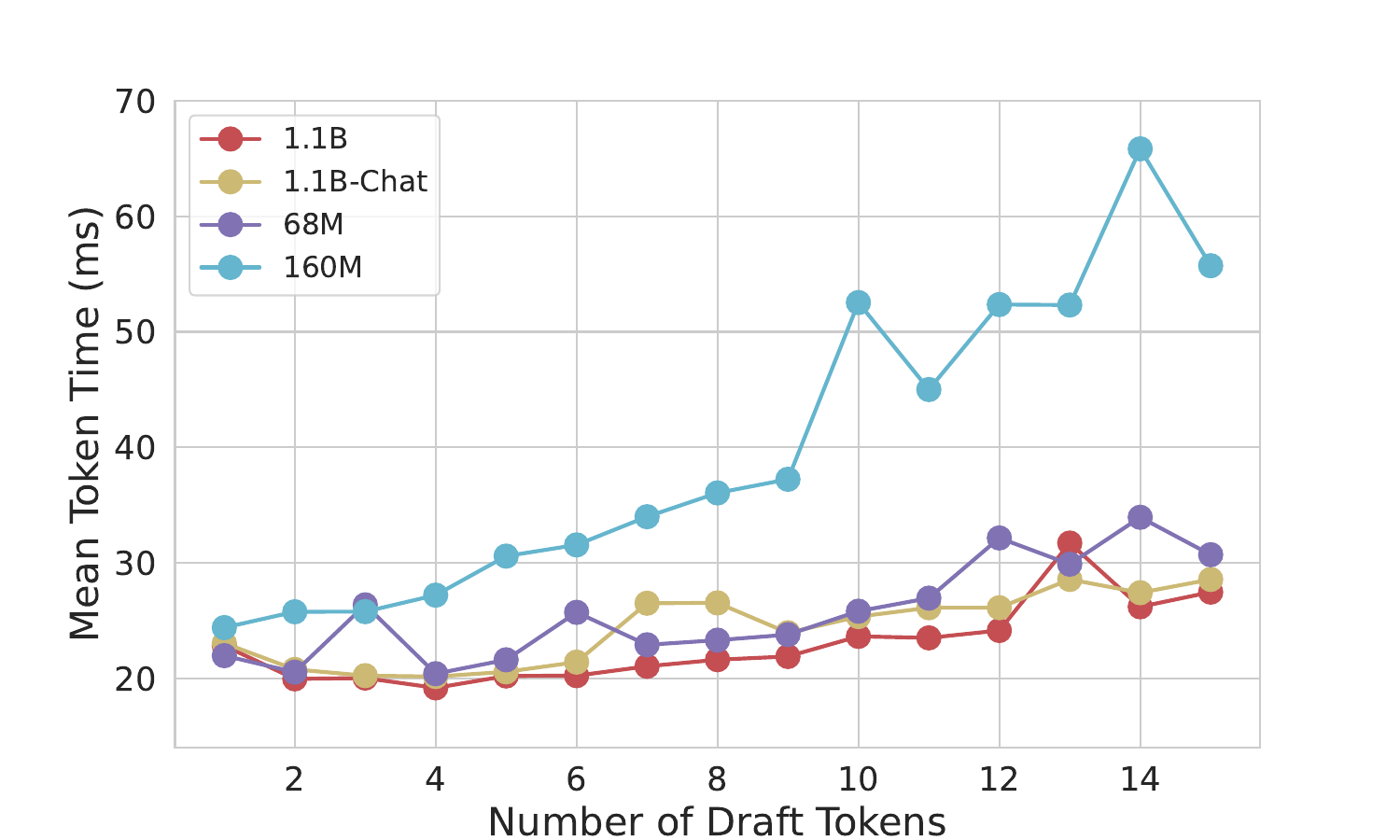}
  \caption{Generation speed of speculative decoding with different draft LMs and numbers of draft tokens (CodeLlama 7B with nucleus sampling on HumanEval).}
  \label{fig:app_spec_codellama7b_nucleus}
\end{minipage}

\vspace{1em}

\begin{minipage}[b]{0.49\linewidth}
  \centering
  \includegraphics[width=\linewidth]{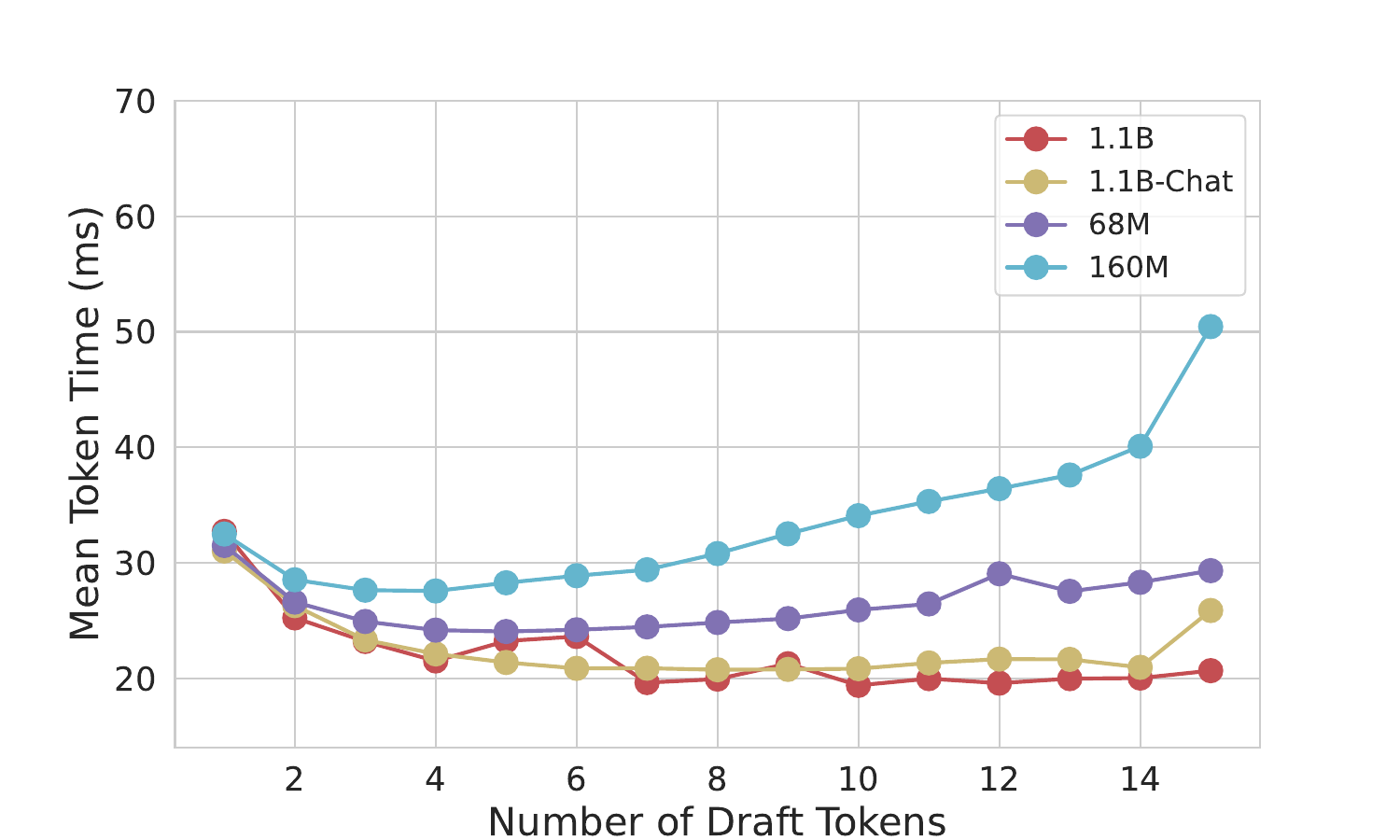}
  \caption{Generation speed of speculative decoding with different draft LMs and numbers of draft tokens (CodeLlama 13B with greedy sampling on HumanEval).}
  \label{fig:app_spec_codellama13b_greedy}
\end{minipage}
\hfill
\begin{minipage}[b]{0.49\linewidth}
  \centering
  \includegraphics[width=\linewidth]{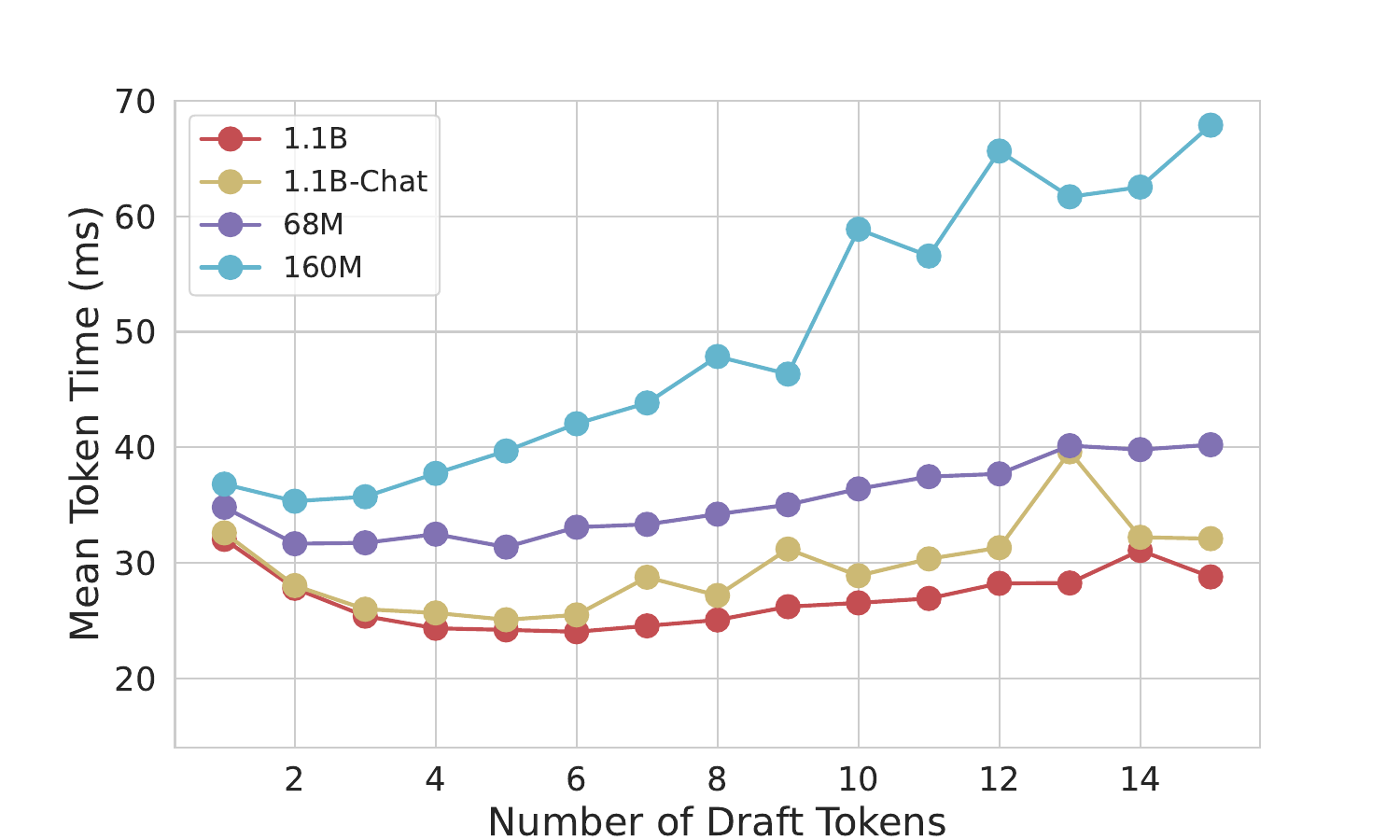}
  \caption{Generation speed of speculative decoding with different draft LMs and numbers of draft tokens (CodeLlama 13B with nucleus sampling on HumanEval).}
  \label{fig:app_spec_codellama13b_nucleus}
\end{minipage}
\end{figure*}

\paragraph{Greedy Sampling} The generation speed of speculative decoding on CodeLlama 7B with greedy sampling is shown in Figure~\ref{fig:app_spec_codellama7b_greedy}. From the figure, we can see that the best setting is to use Llama 68M with 4 draft tokens, resulting in 15.90 ms/token. So, we report 15.90 ms/token in Table~\ref{tab:4_main_results}

\paragraph{Nucleus Sampling} The generation speed of speculative decoding on CodeLlama 7B with nucleus sampling is shown in Figure~\ref{fig:app_spec_codellama7b_nucleus}. From the figure, we can see that the best setting is to use TinyLlama 1.1B with 4 draft tokens, resulting in 18.83 ms/token. So, we report 18.83 ms/token in Table~\ref{tab:4_main_results}.

\subsection{Results on CodeLlama 13B}

\paragraph{Greedy Sampling} The generation speed of speculative decoding on CodeLlama 13B with greedy sampling is shown in Figure~\ref{fig:app_spec_codellama13b_greedy}. From the figure, we can see that the best setting is to use TinyLlama 1.1B with 10 draft tokens, resulting in 19.39 ms/token. So, we report 19.39 ms/token in Table~\ref{tab:4_main_results}

\paragraph{Nucleus Sampling} The generation speed of speculative decoding on CodeLlama 13B with nucleus sampling is shown in Figure~\ref{fig:app_spec_codellama13b_nucleus}. From the figure, we can see that the best setting is to use TinyLlama 1.1B with 6 draft tokens, resulting in 22.68 ms/token. So, we report 22.68 ms/token in Table~\ref{tab:4_main_results}

\subsection{Results on Vicuna 7B}
\begin{figure*}[t]
\centering
\begin{minipage}[b]{0.49\linewidth}
  \centering
  \includegraphics[width=\linewidth]{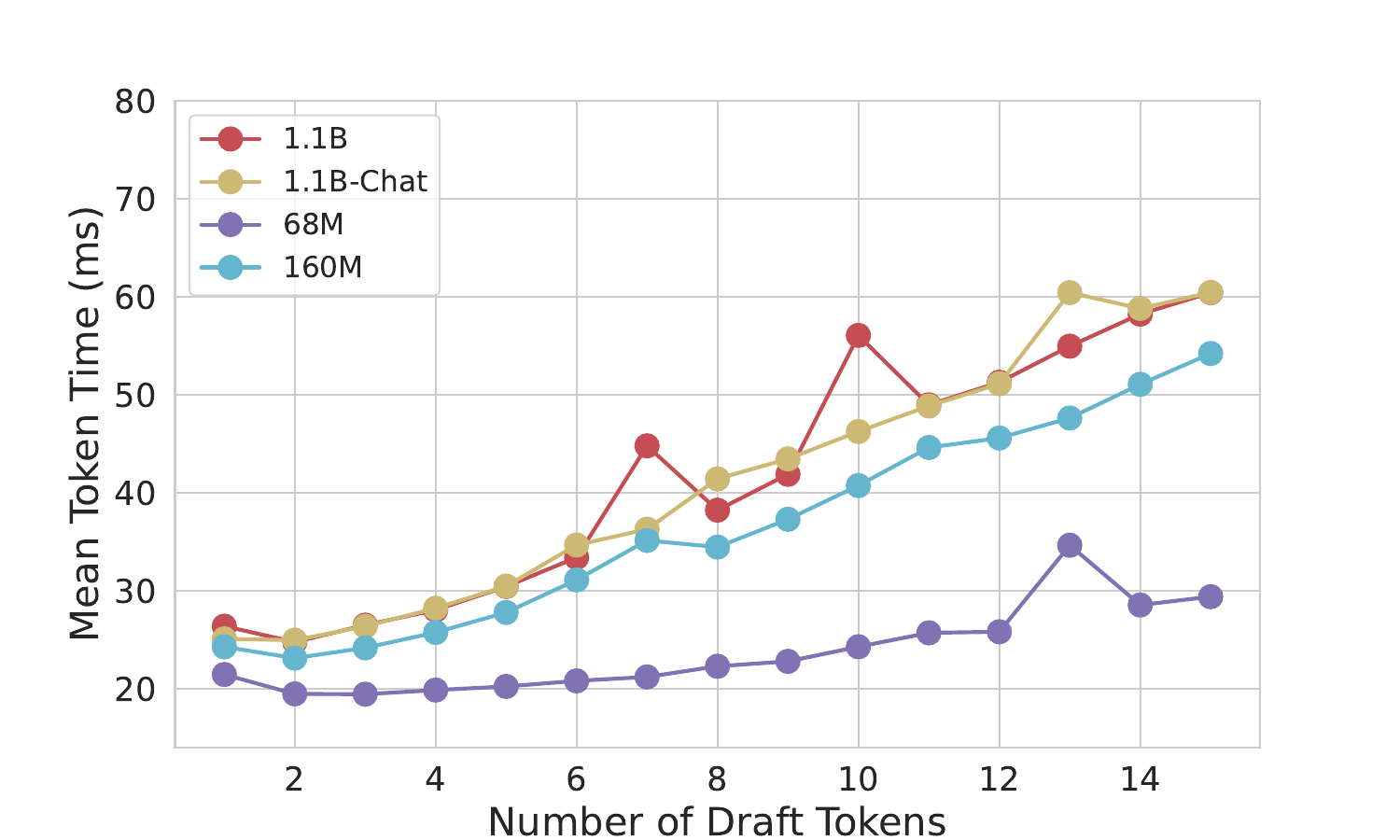}
  \caption{Generation speed of speculative decoding with different draft LMs and numbers of draft tokens (Vicuna 7B with greedy sampling on MT-Bench).}
  \label{fig:app_spec_vicuna7b_greedy}
\end{minipage}
\hfill
\begin{minipage}[b]{0.49\linewidth}
  \centering
  \includegraphics[width=\linewidth]{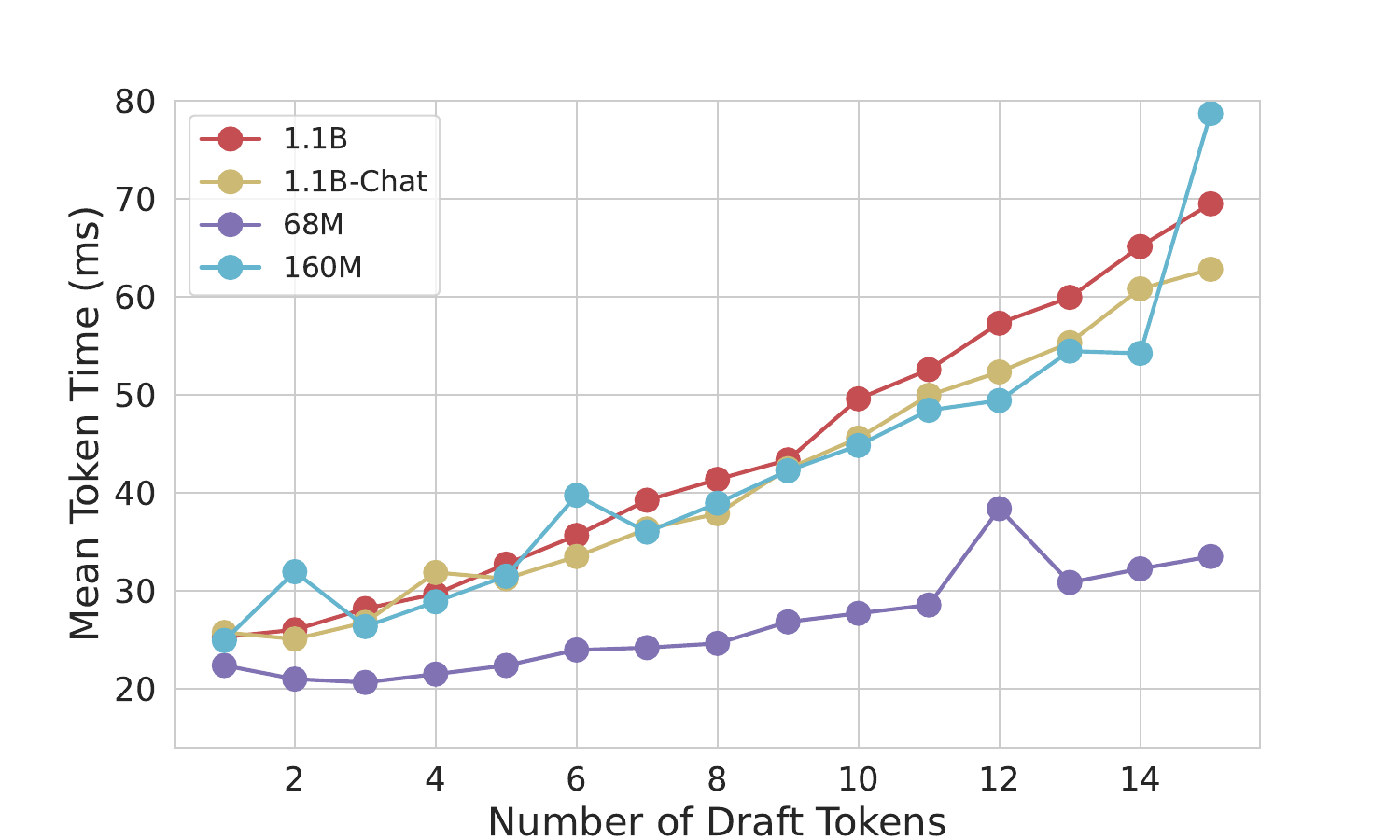}
  \caption{Generation speed of speculative decoding with different draft LMs and numbers of draft tokens (Vicuna 7B with nucleus sampling on MT-Bench).}
  \label{fig:app_spec_vicuna7b_nucleus}
\end{minipage}

\vspace{1em}

\begin{minipage}[b]{0.49\linewidth}
  \centering
  \includegraphics[width=\linewidth]{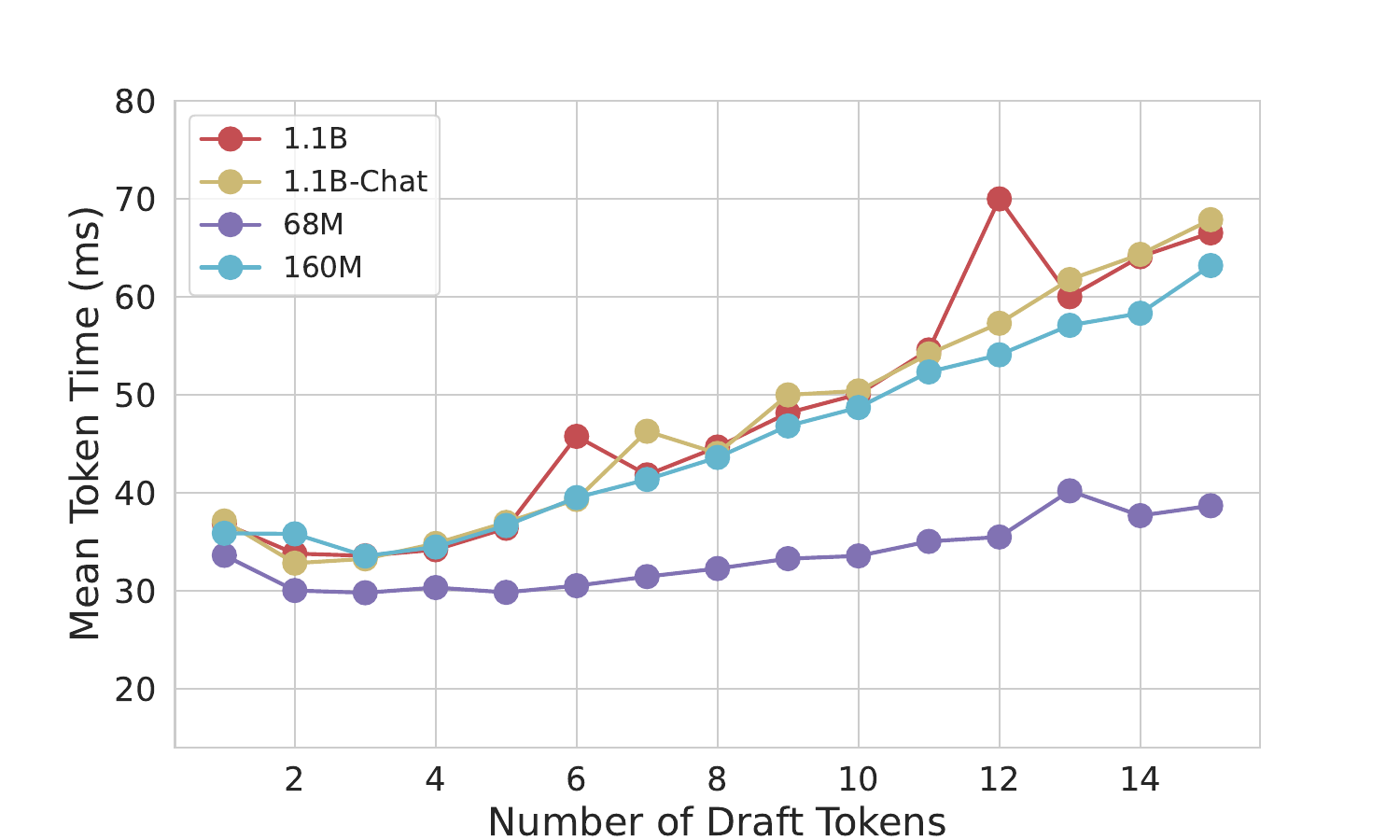}
  \caption{Generation speed of speculative decoding with different draft LMs and numbers of draft tokens (Vicuna 13B with greedy sampling on MT-Bench).}
  \label{fig:app_spec_vicuna13b_greedy}
\end{minipage}
\hfill
\begin{minipage}[b]{0.49\linewidth}
  \centering
  \includegraphics[width=\linewidth]{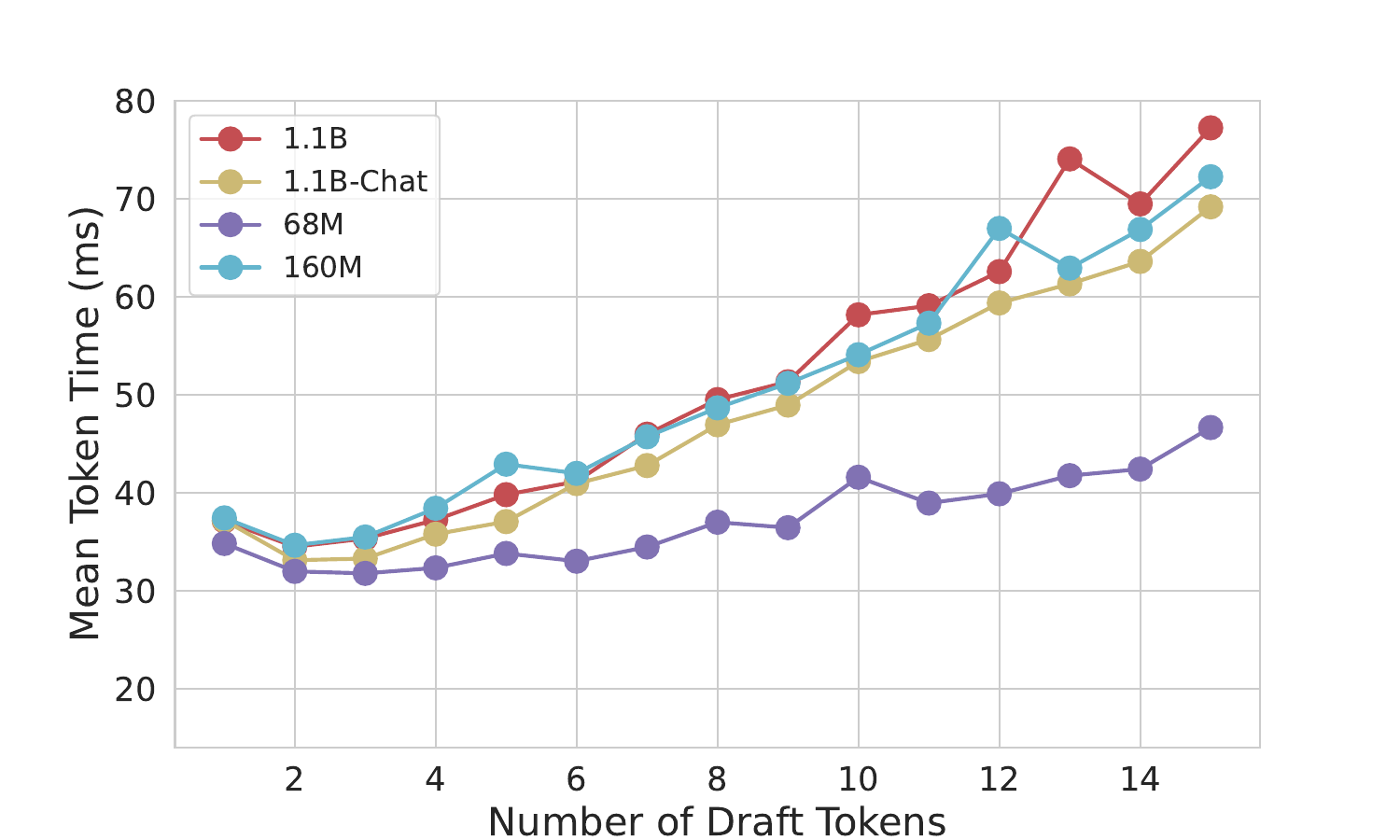}
  \caption{Generation speed of speculative decoding with different draft LMs and numbers of draft tokens (Vicuna 13B with nucleus sampling on MT-Bench).}
  \label{fig:app_spec_vicuna13b_nucleus}
\end{minipage}
\end{figure*}

\paragraph{Greedy Sampling} The generation speed of speculative decoding on Vicuna 7B with greedy sampling is shown in Figure~\ref{fig:app_spec_vicuna7b_greedy}. From the figure, we can see that the best setting is to use Llama 68M with 3 draft tokens, resulting in 19.44 ms/token. So, we report 19.44 ms/token in Table~\ref{tab:4_main_results}

\paragraph{Nucleus Sampling} The generation speed of speculative decoding on Vicuna 7B with nucleus sampling is shown in Figure~\ref{fig:app_spec_vicuna7b_nucleus}. From the figure, we can see that the best setting is to use Llama 68M with 3 draft tokens, resulting in 20.65 ms/token. So, we report 20.65 ms/token in Table~\ref{tab:4_main_results}.

\subsection{Results on Vicuna 13B}

\paragraph{Greedy Sampling} The generation speed of speculative decoding on Vicuna 13B with greedy sampling is shown in Figure~\ref{fig:app_spec_vicuna13b_greedy}. From the figure, we can see that the best setting is to use Llama 68m with 3 draft tokens, resulting in 29.80 ms/token. So, we report 29.80 ms/token in Table~\ref{tab:4_main_results}

\paragraph{Nucleus Sampling} The generation speed of speculative decoding on Vicuna 13B with nucleus sampling is shown in Figure~\ref{fig:app_spec_vicuna13b_nucleus}. From the figure, we can see that the best setting is to use Llama 68M with 3 draft tokens, resulting in 31.78 ms/token. So, we report 31.78 ms/token in Table~\ref{tab:4_main_results}
\section{Additional Ablation Studies}\label{app:ablation}

\begin{table*}[t]
    \centering
    {
    \resizebox{1.9\columnwidth}{!}{
    \vspace{.5em}
    \begin{tabular}{cccccc}
    \toprule
    \textbf{Model} & \textbf{Method} & \textbf{Datastore Size} & \textbf{Retrieval Time} & \textbf{\textit{Mean Token Time}}($\downarrow$) & \textbf{Speedup}($\uparrow$) \\
     \midrule
    \midrule
    Vicuna 7B & Baseline(Greedy) & - & - & 25.48 ms/token & $1\times$\\
     Vicuna 7B & REST(Greedy) & 465 MB & 0.1 ms  &  16.23 ms/token & $1.57 \times$\\
     Vicuna 7B &  REST(Greedy) & 12 GB & 0.6 ms  & 15.12 ms/token & $1.69 \times$\\
     \midrule
    Vicuna 13B & Baseline(Greedy) & - & - & 44.30 ms/token & $1\times$\\
     Vicuna 13B & REST(Greedy) & 465 MB & 0.1 ms  &  27.25 ms/token & $1.63 \times$\\
     Vicuna 13B &  REST(Greedy) & 12 GB & 0.6 ms  & 25.08 ms/token & $1.77 \times$\\
    \bottomrule
    \end{tabular}
    }
    }
    \caption{Generation speed with different datastore sizes with greedy sampling on MT-Bench.}.
    \label{tab:4_abalation_datastore_mtbench_greedy}
\end{table*}

\begin{table*}[t]
    \centering
    {
    \resizebox{1.9\columnwidth}{!}{
    \vspace{.5em}
    \begin{tabular}{cccccc}
    \toprule
    \textbf{Model} & \textbf{Method} & \textbf{Datastore Size} & \textbf{Retrieval Time} & \textbf{\textit{Mean Token Time}}($\downarrow$) & \textbf{Speedup}($\uparrow$) \\
     \midrule
    \midrule
    Vicuna 7B & Baseline(Nucleus) & - & - & 25.93 ms/token & $1\times$\\
     Vicuna 7B & REST(Nucleus) & 465 MB & 0.1 ms  &  16.98 ms/token & $1.53 \times$\\
     Vicuna 7B &  REST(Nucleus) & 12 GB & 0.6 ms  & 16.02 ms/token & $1.62 \times$\\
     \midrule
    Vicuna 13B & Baseline(Nucleus) & - & - & 44.43 ms/token & $1\times$\\
     Vicuna 13B & REST(Nucleus) & 465 MB & 0.1 ms  &  28.00 ms/token & $1.59 \times$\\
     Vicuna 13B &  REST(Nucleus) & 12 GB & 0.6 ms  & 25.92 ms/token & $1.71 \times$\\
    \bottomrule
    \end{tabular}
    }
    }
    \caption{Generation speed with different datastore sizes with nucleus sampling on MT-Bench.}.
    \label{tab:4_abalation_datastore_mtbench_nucleus}
\end{table*}

\begin{figure}[h]
\centering
\includegraphics[width=1.0\linewidth]{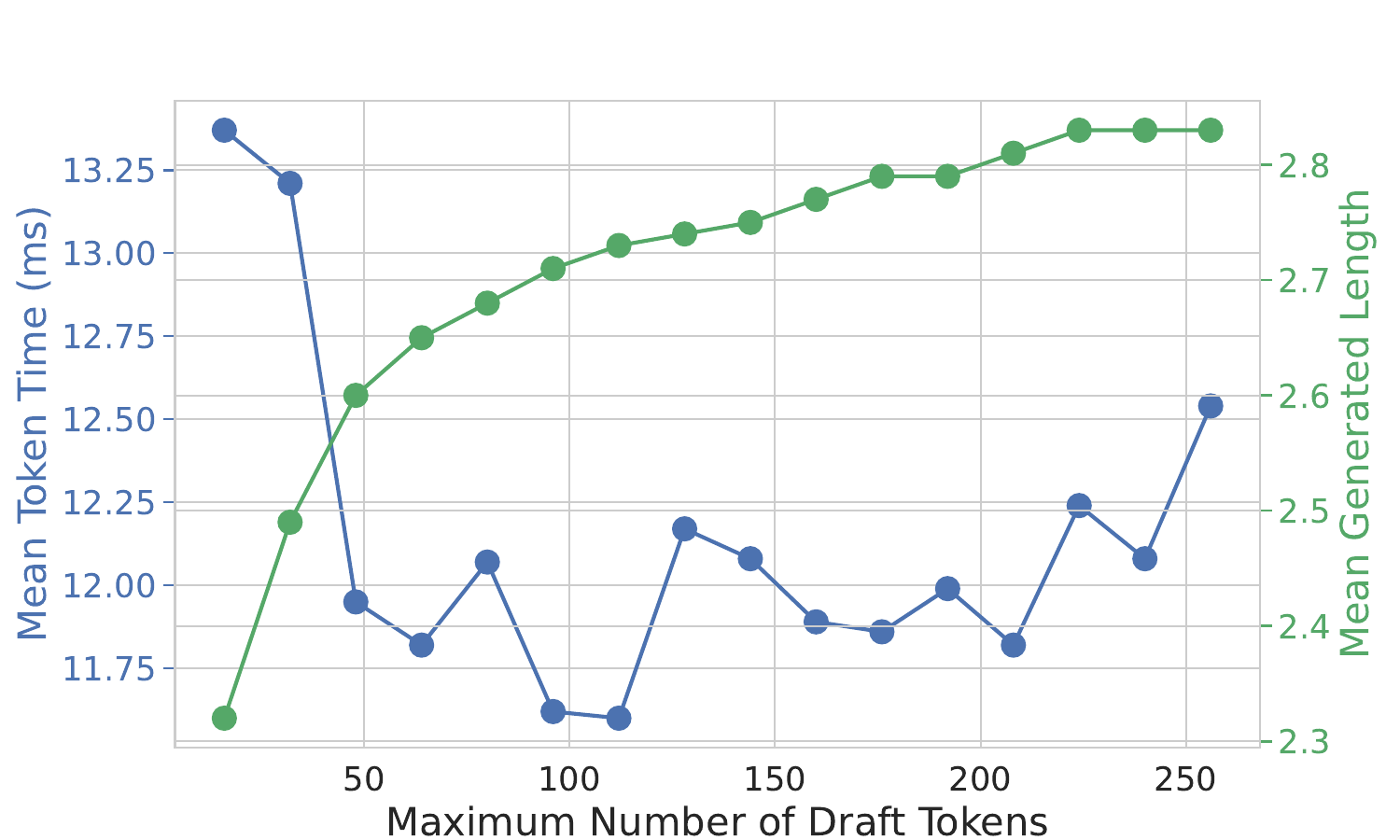}
\vspace{-0.5em}
\caption{
Generation speed of {\ours} with different maximum numbers of selected draft tokens in the Trie (CodeLlama 7B with greedy sampling on the HumanEval). 
}
\label{fig:4_ablation_nodes}
\end{figure}

\paragraph{Effect of the datastore size} For the MT-Bench benchmark, We construct a small datastore with 465 MB derived from ShareGPT\footnote{\url{https://huggingface.co/datasets/Aeala/ShareGPT_Vicuna_unfiltered/blob/main/ShareGPT_2023.05.04v0_Wasteland_Edition.json}} and a large datastore with 12 GB derived from UltraChat~\cite{ultrachat}. In table~\ref{tab:4_abalation_datastore_mtbench_greedy} and table~\ref{tab:4_abalation_datastore_mtbench_nucleus}, we can see that using a larger datastore brings better speedup rates. Moreover, although using a quite small datastore (465MB), REST still achieves a notable speedup.

\paragraph{Effect of the maximum number of draft tokens} Increasing the volume of draft tokens can potentially lead to a higher \textit{Mean Generated Length} by the LLM. However, this also escalates the computational burden on GPUs during verification. As shown in Figure~\ref{fig:4_ablation_nodes}, an initial speed increase is observed as the maximum number of draft tokens increases. However, beyond the threshold of 48 draft tokens, the speed stabilizes to an average of approximately 11.75 ms per token. When the token count exceeds 200, it leads to a slowdown. Therefore, while it is possible to achieve similar speeds with a large maximum number of draft tokens, it's more efficient to limit the number to a smaller one to avoid unnecessary strain on GPUs.

\begin{figure}[h]
\centering
\includegraphics[width=1.0\linewidth]{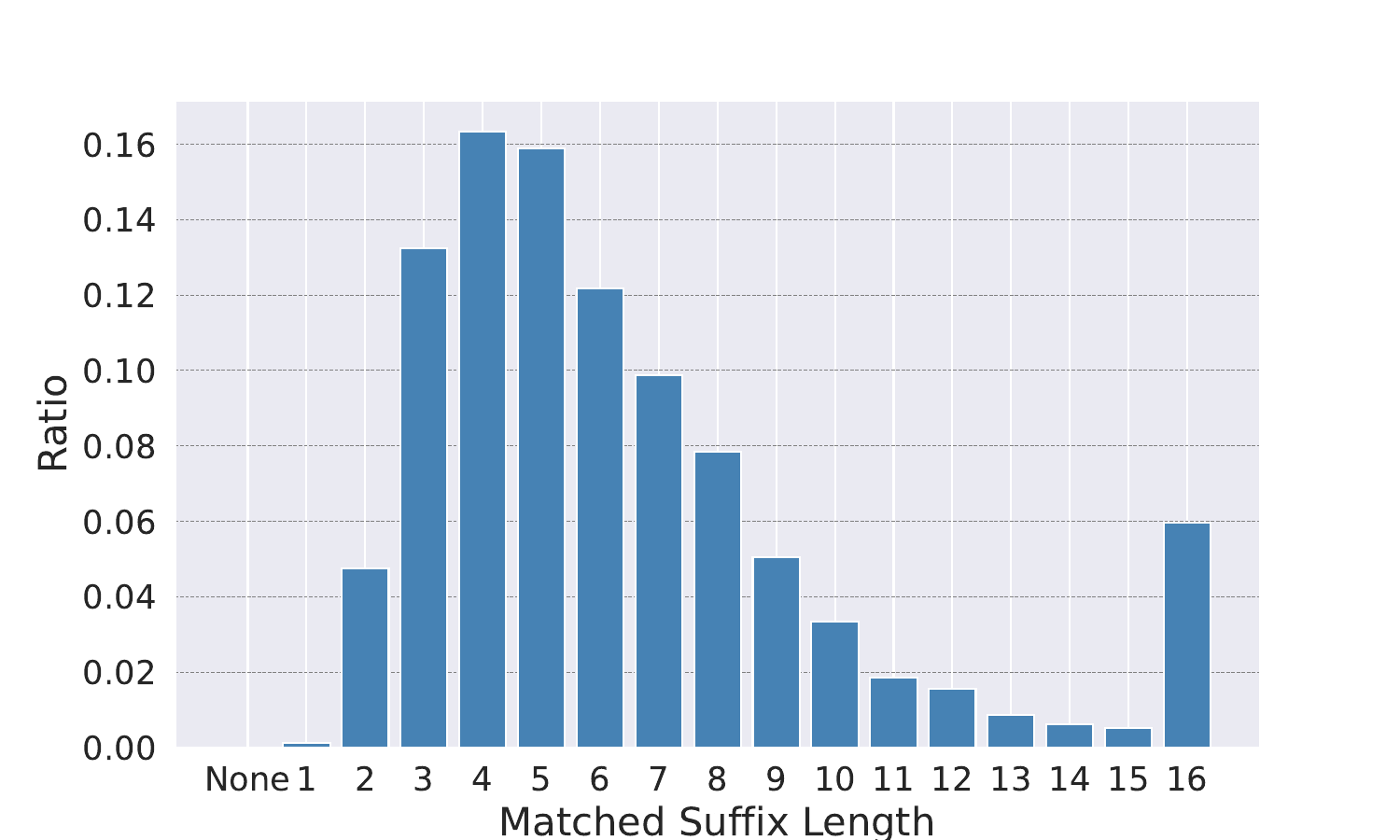}
\vspace{-0.5em}
\caption{
Distribution of matched suffix length (CodeLlama 7B with greedy decoding on HumanEval).
}
\label{fig:4_ablation_matched_span}
\end{figure}
\paragraph{Visualization of matched suffix length}  The distribution of matched suffix length of the context $s$ is illustrated in Figure~\ref{fig:4_ablation_matched_span}.  From this graphic, it is apparent that almost all cases contain a matched suffix length. Notably, shorter suffix lengths ranging from 2 to 9 comprise the majority of the matched cases, accounting for a substantial 85\% of the total. In contrast, longer suffix lengths, which range from 10 to 16, constitute a minority, making up only 15\% of the cases.


\end{document}